\documentclass[sigconf]{aamas}  %
\renewcommand\footnotetextcopyrightpermission[1]{} %
\usepackage{booktabs}

\usepackage{flushend}
\setcopyright{ifaamas}  %
\acmDOI{doi}  %
\acmISBN{}  %
\acmConference[AAMAS'20]{Proc.\@ of the 19th International Conference on Autonomous Agents and Multiagent Systems (AAMAS 2020), B.~An, N.~Yorke-Smith, A.~El~Fallah~Seghrouchni, G.~Sukthankar (eds.)}{May 2020}{Auckland, New Zealand}  %
\acmYear{2020}  %
\copyrightyear{2020}  %
\acmPrice{}  %

\usepackage{placeins}
\usepackage{hyperref}
\usepackage{url}
\usepackage{natbib}
\usepackage{bbm}
\usepackage{caption}
\usepackage{hyperref}       %
\usepackage{url}            %
\usepackage{booktabs}       %
\usepackage{amsfonts}       %
\usepackage{nicefrac}       %
\usepackage{microtype}      %
\usepackage{eucal}
\usepackage{bbm}
\usepackage{amssymb}
\usepackage{verbatim}
\usepackage{amsmath}
\usepackage{subfiles}
\usepackage{graphicx}
\usepackage{natbib}
\usepackage{mathtools}
\usepackage{xcolor}
\usepackage{stfloats}

\definecolor{darkgreen}{RGB}{22, 183, 56}

\newcommand{\method}{CopyCAT}

\DeclareRobustCommand{\patch}{\Delta}

\DeclareRobustCommand{\dataset}{\mathcal{D}}
\DeclareRobustCommand{\adv}{\pi^{\text{target}}}

\newcommand{\norm}[1]{\left\lVert#1\right\rVert_2}
\newcommand{\norminf}[1]{\left\lVert#1\right\rVert_{\infty}}

\DeclareMathOperator*{\argmax}{arg\,max}

\newcommand{\Expect}{\mathop{\mathbb{E}}}
\begin{document}

\title[CopyCAT]{CopyCAT:\\Taking Control of Neural Policies with Constant Attacks}  %

\author{L\'eonard Hussenot}
\affiliation{%
  \institution{Google Research \\INRIA SequeL, Universit\'e de Lille}
}
\author{Matthieu Geist}
\affiliation{%
  \institution{Google Research}
}
\author{Olivier Pietquin}

\affiliation{%
 \institution{Google Research}
}

\begin{abstract}  %
We propose a new perspective on adversarial attacks against deep reinforcement learning agents. Our main contribution is \method{}, a targeted attack able to consistently lure an agent into following an outsider's policy. It is pre-computed, therefore fast inferred, and could thus be usable in a real-time scenario. We show its effectiveness on Atari 2600 games in the novel \textit{read-only} setting. In this setting, the adversary cannot directly modify the agent's \textit{state} --its representation of the environment-- but can only attack the agent's \textit{observation} --its perception of the environment. Directly modifying the agent's state would require a \textit{write-access} to the agent's inner workings and we argue that this assumption is too strong in realistic settings.
\end{abstract}

\maketitle
\thispagestyle{empty}

\section{Introduction}
We are interested in the problem of attacking sequential control systems that use deep neural policies. In the context of supervised learning, previous work developed methods to attack neural classifiers by crafting so-called ``adversarial examples''. These are malicious inputs particularly successful at fooling deep networks with high-dimensional input-data like images. Within the framework of sequential-decision-making, previous works used these adversarial examples only to break neural policies. Yet the attacks they build are rarely applicable in a real-time setting as they require to craft a new adversarial input at each time step. Besides, these methods use the strong assumption of having a \textit{write-access} to what we call the agent's \textit{state}, the actual input of the neural policy built by the algorithm from the observations. When taking this assumption, the adversary --the algorithm attacking the agent-- is not placed at the interface between the agent and the environment where the system is the most vulnerable. We wish to design an attack with a more general purpose than just shattering a neural policy as well as working in a more realistic setting.

Our main contribution is \method{}, an algorithm for taking full-control of neural policies. It produces a simple attack that is: (1)~targeted towards a policy, \textit{i.e.}, it aims at matching a neural policy's behavior with the one of an arbitrary policy; (2)~only altering observation of the environment rather than complete agent's state; (3)~composed of a finite set of pre-computed state-independent masks. This way it requires no additional time at inference hence it could be usable in a real-time setting.

We introduce \method{} in the white-box scenario (Sec.~\ref{copycat_attack}), with \textit{read-only} access to the weights and the architecture of the neural policy. This is a realistic setting as prior work showed that after training substitute models, one could transfer an attack computed on these to the inaccessible attacked model \citep{papernot2016transferability}.   The context is the following: (1)~We are given any agent using a neural-network for decision-making (\textit{e.g.}, the Q-network for value-based agents, the policy network for actor-critic or imitation learning methods) and a \textbf{target policy} we want the agent to follow. (2)~The only thing one can alter is the observation the agent receives from the environment and \textbf{not the full input} of the neural controller (the state). In other words, we are granted a \textit{read-only} access to the agent's inner workings. In the case of Atari 2600 games, the agents builds its state by stacking the last four observations. Attacking the agent's state means writing in the agent's \textit{memory} of the last observations. (3)~The computed attack should be inferred fast enough to be used in \textbf{real-time}.

We stress the fact that targeting a policy is a more general scheme than untargeted attacks where the goal is to stop the agent from taking its preferred action (hoping for it to take the worst). It is also more general than the targeted scheme of previous works where one wants the agent to take its least preferred action or to reach a specific state. In our setting, one can either hard-code or train a target policy. This policy could be minimizing the agent's true reward but also maximizing the reward for another task. For instance, this could mean taking full control of an autonomous vehicle, possibly bringing it to any place of your choice.

We exemplify this approach on the classical benchmark of Atari 2600 games (Sec.\ref{sec:atari_experiments}). We show that taking control of a trained deep RL agent so that its behavior matches a desired policy can be done with this very simple attack. We believe such an attack reveals the vulnerability of autonomous agents. As one could lure them into following catastrophic behaviors, autonomous cars, robots or any agent with high dimensional inputs are exposed to such manipulation. This suggests that it would be worth studying new defense mechanisms that could be specific to RL agents, but this is out of the scope of this paper. 

Section~\ref{additional_experiments} presents experiments studying various aspects of the proposed method, as the use of \method{} in the black-box setting or in environments with a more complex perception. These results assess the possibility of using the proposed method in various settings even though some of these experiments are still preliminary.

\section{Background}\label{background}
In \textbf{Reinforcement Learning} (RL), an agent interacts sequentially with a dynamic environment so as to learn an optimal control. To do so, the problem is modeled as a Markov Decision Process. It is a tuple $\{\mathcal{S},\mathcal{A},P,r,\gamma\}$ with $\mathcal{S}$ the state space, $\mathcal{A}$ the action space we consider as finite in the present work, $P$ the transition kernel defining the dynamics of the environment, $r$ a bounded reward function and $\gamma\in(0,1)$ a discount factor. The policy $\pi$ maps states to distributions over actions: $\pi(\cdot|s)$. The (random) discounted return is defined as $G = \sum_{t \geq 0} \gamma^t r_t$. The policy $\pi$ is trained to maximize the agent expected discounted return. The function $V^\pi(s)=\mathbb{E}_\pi[G|s_0=s]$ denotes the value function of policy $\pi$ (where $\mathbb{E}_\pi[\cdot]$ denotes the expectation over all possible trajectories generated by policy $\pi$). We also call $\mu_0$ the initial state distribution and $\rho(\pi)=\mathbb{E}_{s\sim\mu_0}[V^\pi(s)]$ the expected cumulative reward starting from $\mu_0$.
Value-based algorithms \citep{mnih2015human, hessel2018rainbow} use the value function, or more frequently the action-value function $Q^\pi(s,a)=\mathbb{E}_\pi[G|s_0=s, a_0=a]$, to compute $\pi$.
To handle large state spaces, deep RL uses deep neural networks for function approximation. For instance, value-based deep RL parametrizes the action-value function $Q_{\omega}$ with a neural network of parameters $\omega$ and deep actor-critics \citep{mnih2016asynchronous} directly parametrize their policy $\pi_{\theta}$ with a neural network of parameters $\theta$. In both cases, the taken action is inferred by a forward-pass in a neural network.

\textbf{Adversarial examples} were introduced in the context of supervised classification.
Given a classifier $C$, an input $x$, a bound $\epsilon$ on a norm $\|.\|$, an adversarial example is an input $x'=x+\eta$ such that $C(x) \neq C(x')$ while $\|x-x'\| \leq \epsilon$.
\textit{Fast Gradient Sign Method} (FGSM)~\citep{goodfellow2015explaining} is a widespread method for generating adversarial examples for the $L_{\infty}$~norm. From a linear approximation of $C$, it computes the attack $\eta$ as:
\begin{equation}\label{fgsm}
   \eta = \epsilon \cdot \text{sign}(\nabla_x l(\theta,x,y)),
\end{equation}
with $l(\theta,x,y)$ the loss of the classifier and $y$ the true label.

As an adversary, one wishes to \textbf{maximize} the loss $l(\theta,x+\eta,y)$ w.r.t. $\eta$. Presented this way, it is an \textit{untargeted} attack. It pushes $C$ towards misclassifying $x'$ in \textit{any} other label than $y$.
It can easily be turned into a \textit{targeted} attack by, instead of $l(\theta,x+\eta,y)$,  optimizing for $-l(\theta,x+\eta,y_\text{target})$  with $y_\text{target}$ the label the adversary wants $C$ to predict for $x'$.
This attack, optimized for the $L_{\infty}$ norm can also be turned into an $L_2$ attack by taking:
\begin{equation}\label{fgsm2}
   \eta = \epsilon \cdot \frac{\nabla_x l(\theta,x,y)}{\norm{\nabla_x l(\theta,x,y)}}.
\end{equation}

As shown by Eqs.~\eqref{fgsm} and~\eqref{fgsm2}, these attacks are computed with one single step of gradient, hence the term ``fast''. These two attacks can be turned into --more efficient, yet slower-- iterative methods \citep{carlini2017towards, dong2018boosting} by taking several successive steps of gradients. These methods will be referred to as \textit{iterative-FGSM}.

When using \textbf{deep networks} to compute its policy, an RL agent can be fooled the same way as a supervised classifier. As a policy can be seen as a mapping $\mathcal{S}\to \mathcal{A}$, untargeted FGSM~(\ref{fgsm}) can be applied to a deep RL agent to stop it from taking its preferred action:  $a^*= \argmax_{a \in \mathcal A} \pi(a|s)$. Similarly targeted FGSM can be used to lure the agent into taking a specific action. Yet, this would mean having to compute a new attack at each time step, which is generally not feasible in a real-time setting. Moreover, with this formulation, it needs to directly modify the agent's state, the input of the neural policy, which is a strong assumption.
\section{The CopyCAT attack}\label{copycat_attack}
In this work, we propose \method{}. It is an attack whose goal is to lure an agent into having a given behavior, the latter being specified by another policy. \method{}'s goal is not only to lure the agent into taking specific actions but to fully control its behavior. Formally, \method{} is composed of a set of additive masks $\patch=\{\delta_a\}_{a\in \mathcal{A}}$ that can be used to drive a policy $\pi$ to follow any policy $\adv$. Each additive mask $\delta_a$ is pre-computed to lure $\pi$ into taking a specific action $a$ when added to the current observation regardless of the content of the observation. It is, in this sense, a \textit{universal} attack. \method{} is an attack on raw observations and, as $\patch$ is pre-computed, it can be used online in a real-time setting with no additional computation.

\paragraph{\textbf{Notations}} We denote $\pi$ the attacked policy and $\adv$ the target policy. At time step $t$, the policy $\pi$ outputs an action $a_t$ taking the state $s_t$ as input. The agent state is internally computed from the past observations and we denote $f$ the observations-to-state function: $s_t = f(o_t, o_{1:t-1})$ with $ o_{1:t-1} = (o_{1},  o_{2} ...  o_{t-1})$.

\paragraph{\textbf{Data Collection}} In order to be pre-computed, \method{} needs to gather data from the agent. By watching the agent interacting with the environment, \method{} gathers a dataset $\dataset$ of $K$ episodes made of observations: $\dataset=(o_t^k)^{k\in (1:K)}_{t\in(1,T_k)}$. We recall that the objective in this setting is for \method{} to work with a \textit{read-only} access to the inner workings of the agent. We thus stress that $\dataset$ is made of observations rather than states. If \method{} is successful, $\pi$ is going to behave as $\adv$ and thus may experience observations out of the distribution represented in $\dataset$. Yet, as will be shown, \method{} transfers to unseen observations. We hypothesize that, as we build a \textit{universal} attack, the learned attack is able to move the whole support of observations in a region of $\mathbb{R}^N$ where $\pi$ chooses a precise action.

\paragraph{\textbf{Training}}
A natural strategy for building an adversarial example targeted towards label $\hat{y}$ is the following. Given a classifier $\mathbb{P}(y|x)$ parametrized with a neural network and an input example $x$, one computes the adversarial example $\hat{x}=x+\delta$ by maximizing $\log{\mathbb{P}(\hat{y}|\hat{x})}$ subject to the constraint: $\norminf{\delta} = \norminf{x-\hat{x}} \leq \epsilon$. The adversary then performs either one step of gradient (FGSM) or uses an iterative method \citep{kurakin2016adversarial} to solve the optimization problem.

Instead, \method{} is built for its masks to be working whatever the observation it is applied to. For each $a \in \mathcal{A}$ we build $\delta_a$, the additional masks luring $\pi$ into taking action $a$, by \textbf{maximizing} over $\delta_a$:
\begin{equation}\label{eq:method}
\Expect_{o_t^k \in \dataset} \big[ \log \pi(a|f(o_t^k+\delta_a,  o^k_{1:t-1})) + \alpha \norm{\delta_a} \big] \text{  s.t. }
\norminf{\delta_a} < \epsilon \, .
\end{equation}
We restrict the method to the case where $f$, the function building the agent's inner state from the observations, is differentiable. The Eq.~\ref{eq:method} can be optimized by alternating between gradient steps with adaptive learning rate \citep{kingma2014adam} on mini-batches and projection steps onto the $L_{\infty}$-ball of radius~$\epsilon$. Unlike FGSM, \method{} is a full optimization method. It does not take only one single step of gradient.

\method{} has two main parameters: $\epsilon \in \mathbb{R}^+$, a hard constraint on the $L_{\infty}$~norm of the attack and $\alpha \in \mathbb{R}^+$, a regularization parameter on the $L_2$~norm of the attack.
\paragraph{\textbf{Inference}} Once $\patch$ is computed, the attack can be used on $\pi$ to make it follow any policy $\adv$. At each time step $t$ and given past observations, $\adv$ infers an action $a^{\text{target}}_t$. The corresponding mask $\delta_{a^{\text{target}}_t} \in \patch$ is applied to the last observation $o_t$ before being passed to the agent. No optimization is made at inference and \method{} can thus be used in a setting where the adversary is not let unlimited time to compute an attack between two time-steps.
\clearpage
\section{Related Work}\label{related_work}
 Vulnerabilities of neural classifiers were highlighted by \citet{szegedy2013intriguing} and several methods were developed to create the so-called \textit{adversarial examples}, maliciously crafted inputs fooling deep networks. In sequential-decision-making, previous works use them to attack deep reinforcement learning agents. However these attacks are not always realistic. The method from \citet{huang2017adversarial} uses \textit{fast-gradient-sign method} \citep{goodfellow2015explaining} for the sole purpose of destroying the agent's performance. What's more, it has to craft a new attack at each time step. This implies back-propagating through the agent's network, which is not feasible in real-time. Moreover, it modifies directly the state of the agent by writing in its memory, which is a strong assumption to take on what component of the agent can be altered. The approach of \citet{lin2017tactics} allows the number of attacked states to be divided by four, yet it uses the heavy optimization scheme from \citet{carlini2017towards} to craft their adversarial examples. This is, in general, not doable in a real-time setting. They also take the same strong assumption of having a ``read \& write-access'' to the agent's inner workings. To the best of our knowledge, they are the first to introduce a targeted attack. However, the setting is restricted to targeting one ``dangerous state''. \citet{pattanaik2018robust} proposes a method to lure the agent into taking its least preferred action in order to reduce its performance but still uses computationally heavy iterative methods at each time step.
\citet{pinto2017robust} proposed an adversarial method for robust training of agents but only considered attacks on the dynamic of the environment, not on the visual perception of the agent. \citet{zhang2018learning} and \citet{ruderman2018uncovering} developed adversarial environment generation to study agent's generalization and worst-case scenarios. Those are different from this present work where we enlighten how an adversary might take control of a neural policy.

\section{Atari Experiments}
\label{sec:atari_experiments}
We wish to build an attack targeted towards the policy $\adv$. At a time step $t$, the attack is said to be successful if $\pi$ under attack indeed chooses the targeted action selected by $\adv$. When $\pi$ is not attacked, the attack success rate corresponds to the \textit{agreement rate} between $\pi$ and $\adv$, measuring how often the policies agree along an unattacked trajectory of $\pi$.

Note that we only deal with trained policies and no learning of neural policies is involved. In other words, $\pi$ and $\adv$ are trained and frozen policies.

What we really want to test is the ability of \method{} to lure $\pi$ into having a specific behavior. For this reason, measuring the attack success rate is not enough. Having a high success rate does not necessarily mean the macroscopic behavior of the attacked agent matches the desired one as will be shown further in this section.

\paragraph{\textbf{Cumulative reward as a proxy for behavior}} We design the following setup. The agent has a policy $\pi$ trained with DQN \citep{mnih2015human}. The policy $\adv$ is trained with Rainbow \citep{hessel2018rainbow}. We select Atari games \citep{bellemare2013arcade} with a clear difference in terms of performance between the two algorithms (where Rainbow obtains higher average cumulative reward than DQN).
This way, in addition to measuring the attack success rate, we can compare the cumulative reward obtained by $\pi$ under attack $\rho(\pi)$ to $\rho(\adv)$ as a proxy of how well $\pi$'s behavior is matching the behavior induced by $\adv$. In this setup, if the attacked policy indeed gets cumulative rewards as high as the ones obtained by $\adv$, it will mean that we did not simply turned some actions into other actions we targeted, but that the whole behavior induced by $\pi$ under attack matches the one induced by $\adv$. This idea that, in reinforcement learning, cumulative reward is the right way to monitor an agent's behavior has been used and developed by the inverse reinforcement learning literature. \citet{ng2000algorithms} argued that the value of a policy, \textit{i.e.} its cumulative reward, is the most compact, robust and transferable description of its induced behavior. We argue that measuring cumulative reward is thus a reasonable proxy for monitoring the behavior of $\pi$ under attack. At this point, we would like to carefully defuse a possible misunderstanding. Our goal is not to show that DQN's performance can be improved by being attacked. We simply want to show that its behavior can be fully manipulated by an opponent and we use the obtained cumulative reward as a proxy for the behavior under attack.

\paragraph{\textbf{Baseline}} We set the objective of building a real-time targeted attack. We thus need to compare our algorithm to baselines applicable to this scenario. The fastest state-of-the-art targeted method can be seen as a variation of \citet{huang2017adversarial}. It applies \textit{targeted FGSM} at each time step $t$ to compute a new attack. It first infers the action $a^{\text{target}}$ and then back-propagates through the attacked network to compute their attack. \method{} only infers $a^{\text{target}}$ and then applies the corresponding pre-computed mask. Both methods can thus be considered usable in real-time yet \method{} is still faster at inference.
We set the objective of attacking only observations rather than complete states so we do not need a \textit{write-access} to the agent's inner workings. DQN stacks four consecutive frames to build its inner state. We thus compare \method{} to a version of the method from \citet{huang2017adversarial} where the gradient inducing the FGSM attack is only computed w.r.t the last observation, so it produces an attack comparable to \method{}, \textit{i.e.}, on a single observation. The gradient from Eq.~\ref{fgsm}: $\nabla_{s_t} l(\theta,s_t,a^{\text{target}})$ becomes $\nabla_{o_t} l(\theta,f(o_t, o_{1:t-1}),a^{\text{target}})$. To keep the comparison fair, $a^{\text{target}}$ is always computed with the exact same policy $\adv$ as in \method{}. The policy $\adv : \mathcal{S}\to\mathcal{A}$ is fixed.

FGSM-$L_{\infty}$ has the same parameter $\epsilon$ as \method{}, bounding the $L_{\infty}$~norm of the attack. \method{} has an additional regularization parameter $\alpha$ allowing the attack to have, for a same $\epsilon$, a lower energy and thus be less detectable. We will compare \method{} to the attack from \citet{huang2017adversarial} showing how behaviors of $\pi$ under attacks match $\adv$ when these attacks are of equal energy.

Full optimization-based attacks would not be inferred fast enough to be used in a sequential decision making problem at each time step.

\begin{figure}[htb]
    \centering
    \begin{minipage}{\linewidth}
        \centering
        \includegraphics[width=.8\linewidth]{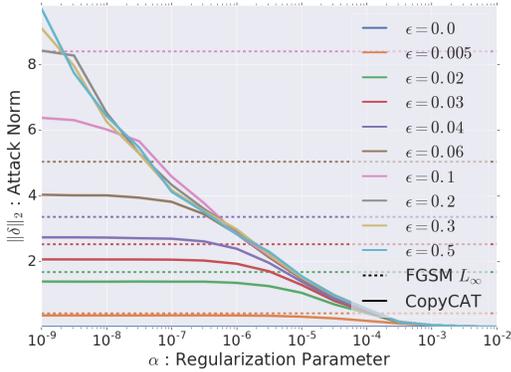}
    \end{minipage}%
\caption{Influence of parameters $\epsilon$ and $\alpha$ on the maximal $L_2$~norm of $\patch$:  $\max_{a\in \mathcal{A}}\norm{\delta_a}$. \method{} is attacking DQN on HERO.}
\label{aed}
\end{figure}

\paragraph{\textbf{Experimental setup}}
We always turn the sticky actions on, which make the problem stochastic \citep{machado2018revisiting}. An attacked observation is always clipped to the valid image range, 0 to 255. For Atari games, DQN uses as its inner state a stack of four observations: $f(o_i,o_{1:i-1})=[o_i, ..., o_{i-3}]$. For learning the masks of $\patch$, we gather trajectories generated by $\pi$ in order to fill $\dataset$ with 10k observations. We use a batch size of 8 and the Adam optimizer \citep{kingma2014adam} with a learning rate of $0.05$. Each point of each plot is the average result over 5 policy $\pi$ seeds and 80 runs for each seed. Only one seed is used for $\adv$ to keep comparison in terms of cumulative reward fair.
\paragraph{\textbf{Attack energy}} As \method{} has an extra parameter $\alpha$, we test its influence on the $L_2$~norm of the produced attack. For a given $\epsilon$, FGSM-$L_{\infty}$ computes an attack $\eta$ of maximal energy. As given by Eq.~\ref{fgsm}, its $L_2$~norm is $\norm{\eta} = \sqrt{N\epsilon^2}$ with $N$ the input dimension. For a given $\epsilon$, \method{} produces $|\mathcal{A}|$ masks. We show in Fig.~\ref{aed} the largest $L_2$~norm of the $|\mathcal{A}|$ masks for a varying $\alpha$ (plain curves) and compare it to the norm of the FGSM-$L_{\infty}$ attack (dashed lines). We want to stress that the attacks are agnostic to the training algorithm so the results are easily transferred to other agents using neural policies trained with another algorithm.

As can be seen on Fig.~\ref{aed}, for a given $\epsilon$ and for the range of tested $\alpha$, the attack produced by \method{} has lower energy than FGSM-$L_{\infty}$. This is especially significant for higher values of $\epsilon$, \textit{e.g} higher than 0.05.

\paragraph{\textbf{Influence of parameters over the resulting behavior}} We wish to show how the agent behaves under attack. As explained before, this analysis is twofold. First, we study results in terms of attack success rate --rate of action chosen by $\pi$ matching $a^{\text{target}}$ when shown attacked observations-- as done in supervised learning. Second, we study the behavior matching through the cumulative rewards under attack $\rho(\pi)$.

What we wish to verify in the following experiment is \method{}'s ability to lure an agent into following a specific behavior. If the attack success rate is high (close to 1), we know that, on a supervised-learning perspective, our attack is successful: it lures the agent into taking specific actions. If, in addition, the average cumulative reward obtained by the agent under attack reaches $\rho(\adv)$ it means that the attack is really successful in terms of behavior. We recall that we attack a policy with a target policy reaching higher average cumulative reward.
\begin{figure}[htb]
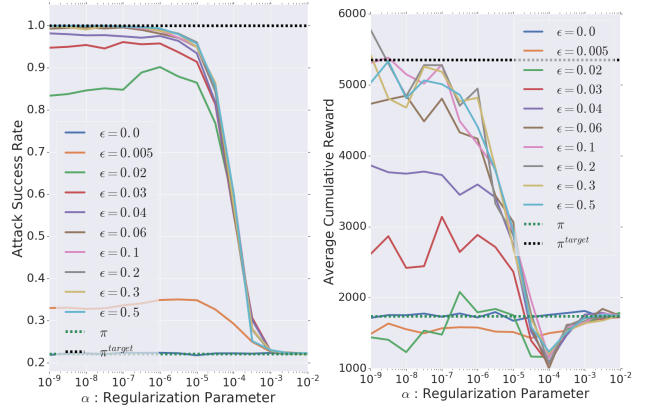

    \centering
    \begin{minipage}{.49\linewidth}
        \centering
        \includegraphics[width=\linewidth]{figures/a2rate_spaceinvaders.pdf}
    \end{minipage}%
    \begin{minipage}{.49\linewidth}
        \centering
        \includegraphics[width=\linewidth]{figures/a2reward_spaceinvaders.pdf}
    \end{minipage}
\caption{Influence of parameters $\epsilon$ and $\alpha$ on the average behavior of the agent playing Space Invaders under \method{} attack . On the left, the attack success rate. On the right, the average cumulative reward.}
\label{a2spaceinvaders}
\end{figure}
\begin{figure}[htb]
    \centering
    \begin{minipage}{.49\linewidth}
        \centering
        \includegraphics[width=\linewidth]{figures/a2rate_hero.pdf}
    \end{minipage}%
    \begin{minipage}{.49\linewidth}
        \centering
        \includegraphics[width=\linewidth]{figures/a2reward_hero.pdf}
    \end{minipage}
\caption{Influence of parameters $\epsilon$ and $\alpha$ on the average behavior of the agent playing HERO under \method{} attack . On the left, the attack success rate. On the right, the average cumulative reward.}
\label{a2hero}
\end{figure}
\begin{figure}[htb]
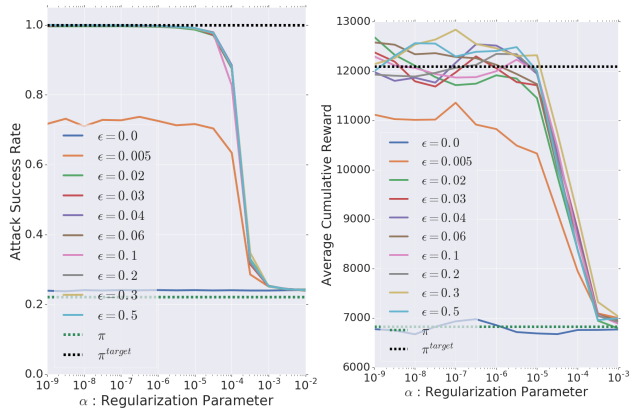

    \centering
    \begin{minipage}{.49\linewidth}
        \centering
        \includegraphics[width=\linewidth]{figures/a2rate_airraid.pdf}
    \end{minipage}%
    \begin{minipage}{.49\linewidth}
        \centering
        \includegraphics[width=\linewidth]{figures/a2reward_airraid.pdf}
    \end{minipage}
\caption{Influence of parameters $\epsilon$ and $\alpha$ on the average behavior of the agent playing Air Raid under \method{} attack . On the left, the attack success rate. On the right, the average cumulative reward.}
\label{a2airraid}
\end{figure}

We show on Fig.~\ref{a2spaceinvaders}, \ref{a2hero} and \ref{a2airraid} (three different games) the attack success rate (left) and the cumulative reward (right) for \method{} (plain curves) for different values of the parameters $\alpha$ and $\epsilon$, as well as for unattacked $\pi$ (green dashed line) and $\adv$ (black dashed lines). We observe  a gap between having a high success rate and forcing the behavior of $\pi$ to match the one of $\adv$. There seems to exist a threshold corresponding to the minimal success rate required for the behaviors to match. For example, as seen on the left, \method{} with $\epsilon=5$ and $\alpha<10^{-5}$ (green curve) is enough to get a 85\% success rate on the attack. However, as seen on the right, it is not enough to get the behavior of $\pi$ under attack to match the one of the target policy as the reward obtained under attack never reaches $\rho(\adv)$.

Overall, we observe on Fig.~\ref{a2spaceinvaders}-right, Fig.~\ref{a2hero}-right and Fig.~\ref{a2airraid}-right that with $\epsilon$ high enough $\epsilon \geq 0.04$ and $\alpha<10^{-6}$, \method{} is able to consistently lure the agent into following the behaviour induced by $\adv$.

\paragraph{\textbf{Comparison to \citet{huang2017adversarial}}} We compare \method{} to the targeted version of FGSM on a setup where the gradient is computed only on the last observation.
As in the last paragraph, we study both the attack success rate and the average cumulative reward under attack. We ask the question: is \method{} able to lure the agent into following the targeted behavior? Is it better at this task than FGSM in the \textit{real-time} and \textit{read-only} setting?
\begin{figure}[htb]
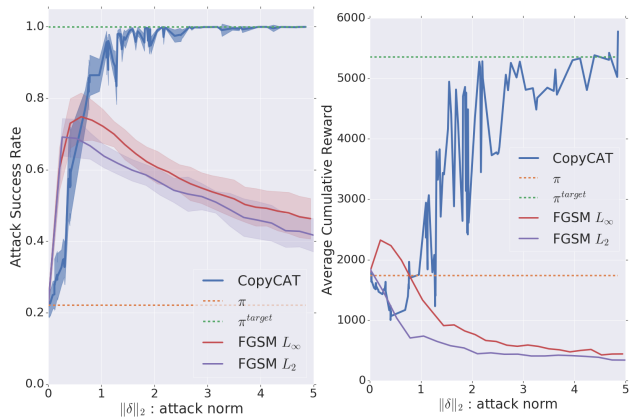

    \centering
    \begin{minipage}{.49\linewidth}
        \centering
        \includegraphics[width=\linewidth]{figures/d2rate_spaceinvaders.pdf}
    \end{minipage}%
    \begin{minipage}{.49\linewidth}
        \centering
        \includegraphics[width=\linewidth]{figures/d2reward_spaceinvaders.pdf}
    \end{minipage}
\caption{\label{d2spaceinvaders}\method{} against FGSM for policy targeted attacks in Space Invaders. On the left, the attack success rate. On the right, the average cumulative reward under attack.}
\end{figure}
\begin{figure}[htb]
    \centering
    \begin{minipage}{.49\linewidth}
        \centering
        \includegraphics[width=\linewidth]{figures/d2rate_hero.pdf}
    \end{minipage}%
    \begin{minipage}{.49\linewidth}
        \centering
        \includegraphics[width=\linewidth]{figures/d2reward_hero.pdf}
    \end{minipage}
\caption{\label{d2hero}\method{} against FGSM for policy targeted attacks in HERO. On the left, the attack success rate. On the right, the average cumulative reward under attack.}
\end{figure}
\begin{figure}[htb]
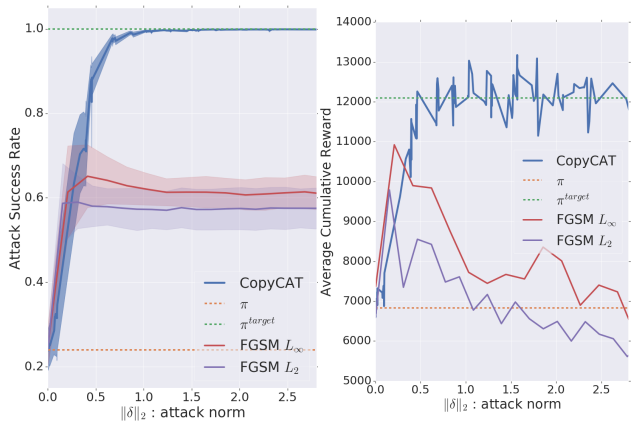

    \centering
    \begin{minipage}{.49\linewidth}
        \centering
        \includegraphics[width=\linewidth]{figures/d2rate_airraid.pdf}
    \end{minipage}%
    \begin{minipage}{.49\linewidth}
        \centering
        \includegraphics[width=\linewidth]{figures/d2reward_airraid.pdf}
    \end{minipage}
\caption{\label{d2airraid}\method{} against FGSM for policy targeted attacks in Air Raid. On the left, the attack success rate. On the right, the average cumulative reward under attack.}
\end{figure}

We show on Fig.~\ref{d2spaceinvaders}, \ref{d2hero} and \ref{d2airraid} (three different games) the success rate of \method{} and FGSM  (y-axis, left) and the average cumulative reward under attack (y-axis, right). These values are plotted (i) against the $L_2$ norm of the attack for FGSM and (ii) against the largest $L_2$ norm of the masks:  $\max_{a\in \mathcal{A}} \norm{\delta_a}$ for \method{}.
We only plot the standard deviation on the attack success rate because it corresponds to the intrinsic noise of \method{}. We do not plot it for cumulative reward for the reason that one seed of $\adv$ has a great variance (with the sticky actions) and matching $\adv$, even perfectly, implies matching the variance of its cumulative rewards. The same phenomenon can be observed on Fig.~\ref{a2spaceinvaders} and~\ref{a2hero}: \method{} is not itself unstable (left figures, when $\alpha$ decreases or $\epsilon$ increases, the rate of successful attacks consistently increases). Yet the cumulative reward is noisier, as the behavior of $\pi$ is now matching with a high-variance policy.
As observed on Fig.~\ref{d2spaceinvaders}-right, Fig.~\ref{d2hero}-left and Fig.~\ref{d2airraid}-left, FGSM is able to turn a potentially significant part of the taken actions into the targeted actions (maximal success rate around 75\% on Space Invaders). However, it is never able to make $\pi$'s behavior match with $\adv$'s behavior as seen on Fig.~\ref{d2spaceinvaders}-right, Fig.~\ref{d2hero}-right and Fig.~\ref{d2airraid}-right. The average cumulative reward obtained by $\pi$ under FGSM attack never reaches the one of $\adv$. On the contrary, \method{} is able to successfully lure $\pi$ into following the desired macroscopic behavior. First, it turns more than 99\% of the taken actions into the targeted actions. Second, it makes $\rho(\pi)$ under attack reach $\rho(\adv)$. Moreover, it does so using only a finite set of masks while the baselines compute a new attack at each time step.

\begin{figure*}[tp]
    \centering
    \begin{minipage}{.8\linewidth}
        \centering

        \includegraphics[width=\linewidth]{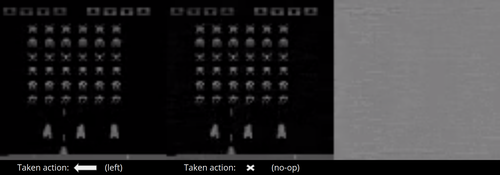} 
    \end{minipage}%
\caption{DQN playing Space Invaders. On the left, the unattacked current observation. In the middle, a mask of \method{} is applied to lure the agent into taking the "no-op" action. Parameters used are: $\epsilon=3\cdot10^{-2}$, $\alpha = 1.3\cdot10^{-5}$. The $L_2$ norm of the corresponding mask (shown on the right) is 0.78. For this same $\epsilon$, FGSM-$L_{\infty}$ would produce an attack of $L_2$ norm: 2.52}
\label{copycat}
\end{figure*}

An example of \method{} is shown on Fig.~\ref{copycat}. The patch $\delta_a$ aiming at action $a$: "no-op" (\textit{i.e.} do nothing) is applied to an agent playing Space Invaders. The patch itself can be seen on the right (gray represents a zero pixel, black negative and white positive). The unattacked observation is on the left, and the attacked one on the right. Below the images is provided the action taken by the same policy $\pi$ when shown the different situations in an online setting. 

\section{Additional experiments}\label{additional_experiments}

In this section, we provide additional experiments to study further various aspects of the proposed approach. We build experiments to assess the possibility that \method{} would be successfully applicable in harder contexts, with very dissimilar policies $\pi$ and $\adv$, in the much more complex black-box setting or in environments with more realistic images than Atari.

\subsection{Towards black-box targeted attacks}
\label{subsec:blackbox}
\citet{papernot2016transferability} observed the transferability of adversarial examples between different models. Thanks to this transferability, one is able to attack a model without having access to its weights. By learning attacks on proxy models, one can build black-box adversarial examples. However \citet{kos2017delving} enlightened the difficulty for the state-of-the-art methods to build \textit{targeted} adversarial examples in this black-box setting. Starting from the intuition that universal attacks may transfer better between models, we enhanced \method{} for it to work in the black-box setting. 

We consider a setting where the adversary (1) is given a set of proxy models $\{\pi_1, ..., \pi_n\}$ trained with the same algorithm as $\pi$, (2) can also query the attacked model $\pi$, but (3) has no access to its weights.
In the black-box setting, \method{} is divided into two steps: first training multiple additional masks, and second selecting the highest performing ones.

\paragraph{\textbf{Training}} The ensemble-based method from \citet{kos2017delving} computes its additional mask by attacking the classifier given by the mean predictions of the proxy models. We instead consider that our attack should be efficient against any convex combination of the proxy models' predictions. For each action $a\in \mathcal{A}$, we compute the mask $\delta_a$ by maximizing over 100 epochs on the dataset $\dataset$:
\begin{equation}
\begin{split}
\Expect_{\substack{o_t^k \in \dataset \\ (\alpha_1...\alpha_p) \sim \Delta}} & \Big[ \log \sum_{1 \leq p \leq n} \alpha_p \pi_{p}(a|f(o_t^k+\delta_a,  o^k_{1:t-1})) + \alpha \norm{\delta_a} \Big] 
\\
& \text{  s.t. }\norminf{\delta_a} < \epsilon \, .
\end{split}
\end{equation}
with $\Delta$ the uniform distribution over the $n$-simplex. By sampling uniformely over the simplex, we hope to build an attack fooling any classifier with predictions in the convex hull of the proxy models' predictions.

For each action, 100 masks are computed this way. These masks are just computed with different random seeds.

\paragraph{\textbf{Selection}} We then compute a \textit{competition accuracy} for each of these random seeds. This accuracy is computed by querying $\pi$ on states built as follows. We take four consecutive observations in $\dataset$, apply 3 masks randomly selected among the previously computed masks on the first 3 observations; the mask $\delta_a$ that is actually being tested is applied on the last observation. The attack is considered successful if $\pi$ outputs the action $a$ corresponding to $\delta_a$.
For each action, the mask with the highest \textit{competition accuracy} among the 100 computed masks is selected.

\paragraph{\textbf{Inference}} The selected masks are then used online as in the white-box setting.

\begin{figure*}[tp]
    \centering
    \includegraphics[width=\linewidth]{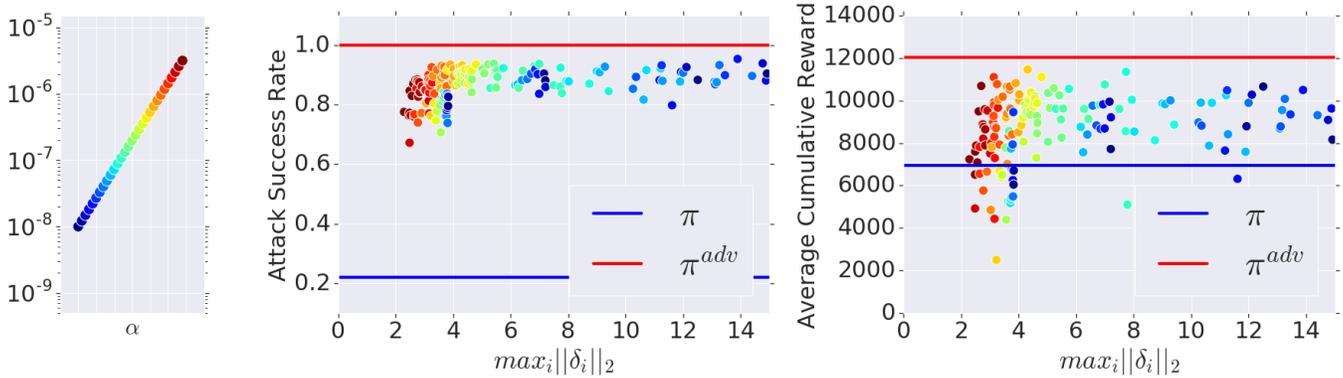}
    \caption{\method{} attacking DQN on Air Raid, in the black-box setting to make it match Rainbow's policy.\label{fig:blackbox}}
\end{figure*}

\paragraph{\textbf{Results}} We provide preliminary results for the considered black-box setting. Four proxy models of DQN are used to attack $\pi$. Again, it is attacked to make it follow the policy $\adv$ given by Rainbow. The results can be found in Fig.~\ref{fig:blackbox}. Each dot is an attack tested over 80 new episodes. Y-axis is the mean success rate (middle) or the cumulative reward (right). X-axis is the maximal norm of the attack. The figure on the left gives the value of $\alpha$ (on the y-axis) corresponding to each color.

We can observe on Fig.~\ref{fig:blackbox} that the proposed black-box attack is effective, even if less efficient than its white-box counterpart. The proposed black-box \method{} could certainly be improved, and we let this for future work.

\subsection{Attacking untrained DQN}
\label{subsec:untrained}

In Sec.~\ref{sec:atari_experiments}, the attacked agent was a trained DQN agent, while the target policy was a trained Rainbow agent. If these agents have clearly different behaviors, one could argue that they were initially trained to solve the same task (getting the highest score as possible), Rainbow achieving better results. 
To further assess \method{}'s ability to lure a policy into following another policy, we therefore attack an untrained DQN, with random weights, to follow the policy $\adv$ still obtained from a trained Rainbow agent. These two policies are dissimilar as one is random while the other is trained to maximized the game's score.
\begin{figure}[!htb]
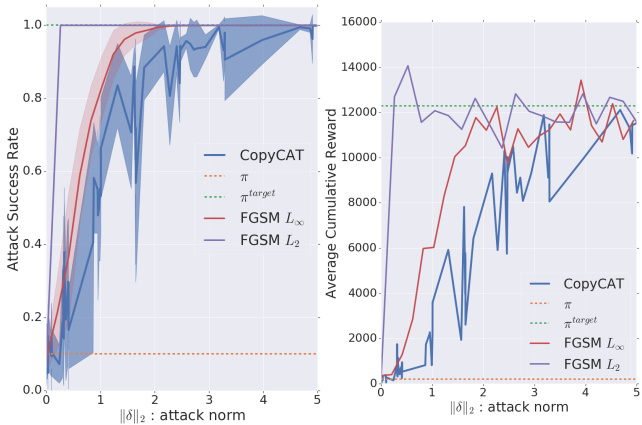

    \centering
    \begin{minipage}{.5\linewidth}
        \centering
        \includegraphics[width=\linewidth]{figures/d2rate_uninit_airraid.pdf}
    \end{minipage}%
    \begin{minipage}{.5\linewidth}
        \centering
        \includegraphics[width=\linewidth]{figures/d2reward_uninit_airraid.pdf}
    \end{minipage}
    \caption{Untrained DQN attacked by \method{} to follow Rainbow's policy on Air Raid. Left: the attack success rate. Right: the average cumulative reward.}
    \label{uninit1}
\end{figure}

\begin{figure}[!htb]
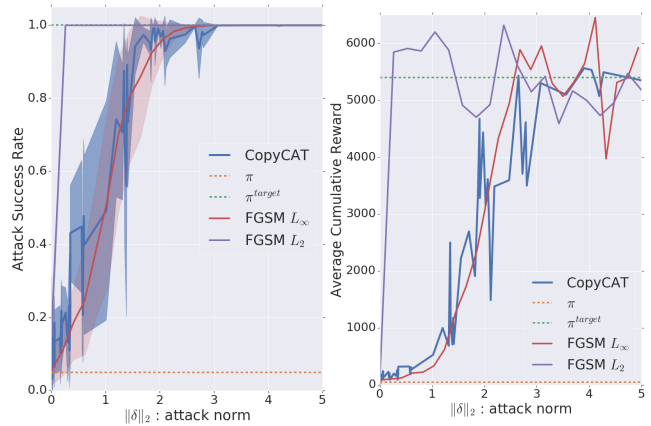

    \centering
    \begin{minipage}{.5\linewidth}
        \centering
        \includegraphics[width=\linewidth]{figures/d2rate_uninit_spaceinvaders.pdf}
    \end{minipage}%
    \begin{minipage}{.5\linewidth}
        \centering
        \includegraphics[width=\linewidth]{figures/d2reward_uninit_spaceinvaders.pdf}
    \end{minipage}
    \caption{Untrained DQN attacked by \method{} to follow Rainbow's policy on Space Invaders. Left: the attack success rate. Right: the average cumulative reward.}
    \label{uninit2}
\end{figure}

\begin{figure}[!htb]
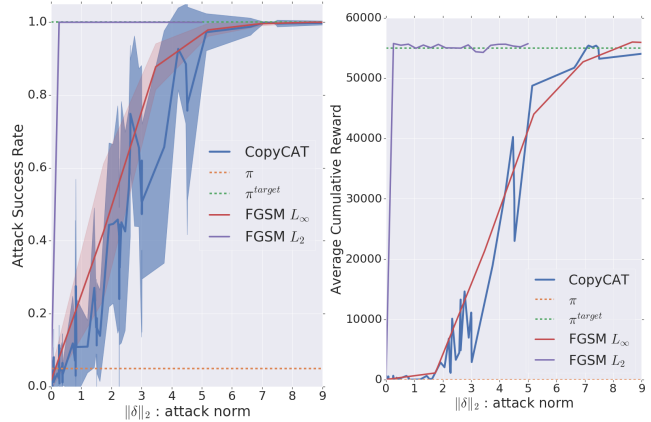

    \centering
    \begin{minipage}{.5\linewidth}
        \centering
        \includegraphics[width=\linewidth]{figures/d2rate_uninit_hero.pdf}
    \end{minipage}%
    \begin{minipage}{.5\linewidth}
        \centering
        \includegraphics[width=\linewidth]{figures/d2reward_uninit_hero.pdf}
    \end{minipage}
\caption{Untrained DQN attacked by \method{} to follow Rainbow's policy on HERO. Left: the attack success rate. Right: the average cumulative reward.}
\label{uninit3}
\end{figure}

We see on Fig.~\ref{uninit1},~\ref{uninit2},~\ref{uninit3} that in this case, FGSM is able to lure $\pi$ into following $\adv$ at least as well as \method{}. This shows that it is easier to fool  an untrained network than a trained one. As expected, trained networks are more robust to adversarial examples. \method{} is also able to lure the agent into following $\adv$.

\subsection{More realistic images}

\label{subsec:perception}
Reinforcement learning led to great improvements for games \citep{silver2017mastering} or robots manipulation \citep{levine2016end} but is not able yet to tackle realistic-image environments.
While this paper focuses on weaknesses of reinforcement learning agents, the relevance of the proposed method would be diminished if one could not compute universal adversarial examples on realistic datasets. We thus present this proof-of-concept, showing the existence of universal adversarial examples on the ImageNet~\citep{imagenet_cvpr09} dataset. Note that \citet{brown2017adversarial} already showed the existence of universal attacks but considered a patch completely covering a part of the image rather than an additional mask on the image.

\begin{figure}[htb]
    \centering
    \begin{minipage}{\linewidth}
        \centering
        \includegraphics[width=0.55\linewidth]{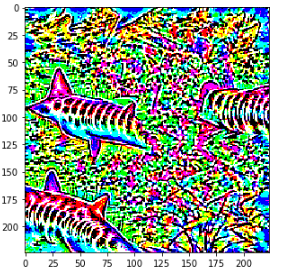}
    \end{minipage}%
        \caption{The rescaled attack for the target class ``tiger\_shark''.\label{tigershark}}
\end{figure}

\begin{figure}[htb]
    \centering
    \begin{minipage}{\linewidth}
        \centering
        \includegraphics[width=\linewidth]{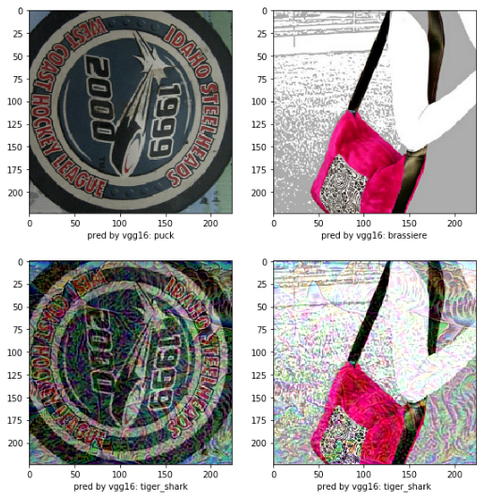}
        \caption{Top: images from the test set, unseen during the training of the attack. Down: attacked images: (image + attack). Below each image, the label predicted by VGG16.\label{imagenet}}
    \end{minipage}%
\end{figure}

We computed a universal attack on VGG16 \citep{simonyan2014very}, targeted towards the label ``tiger\_shark'', the same way \method{} does. It is trained on a small training set (10 batches of size 8), and tested on a random subset of ImageNet validation dataset. The network is taken from Keras pretrained models \citep{chollet2015keras} and attacked in a white-box setting. The same procedure as \method{} is used. 1000 images are randomly selected in the validation set. Only 80 are used for training and the rest is used for testing. The attack is trained with the same loss, same learning rate and same batch size as \method{}, for 200 epochs.
The (rescaled) computed attack is shown in Fig.~\ref{tigershark}. Examples of attacked images from the test set are visible on Fig.~\ref{imagenet}.

After 200 epochs, the train accuracy is 90\% and the test accuracy 88.44\%. This proof-of-concept experiment validates the existence of universal adversarial examples on realistic images and shows that \method{}'s scope is not reduced to Atari-like environments. More generally, the existence of adversarial examples have been shown to be a property of high-dimensional manifolds \citep{goodfellow2015explaining}. Going towards more realistic images, hence higher dimensional images, should on the opposite, allow \method{} to more easily find universal adversarial examples.

\section{Conclusion}

In this work, we built and showed  the effectiveness of \method{}, a simple algorithm designed to attack neural policies in order to manipulate them. We showed its ability to lure a policy into having a desired behavior with a finite set of additive masks, usable in a real-time setting while being applied only on observations of the environment. We demonstrated the effectiveness of these universal masks in Atari games, in the white-box setting. We also investigated an extension of \method{} in the black-box setting and validated the possibility of using it on potentially more complex environments with realistic observations.

As this work shows that one can easily manipulate a policy's behavior, a natural direction of work is to develop robust algorithms, either able to keep their normal behaviors when attacked or to detect attacks to treat them appropriately. Notice however that in a sequential-decision-making setting, detecting an attack is not enough as the agent cannot necessarily stop the process when detecting an attack and may have to keep outputting actions for incoming observations. It is thus an exciting direction of work to develop algorithm that are able to maintain their behavior under such manipulating attacks.
Another interesting direction of work in order to build real-life attacks is to further develop targeted attacks on neural policies in the black-box scenario, with no access to network's weights and architecture. However, targeted adversarial examples are harder to compute than untargeted ones and we may experience more difficulties in reinforcement learning than supervised learning. Indeed, learned representations are known to be less interpretable and the variability between different random seeds to be higher than in supervised learning. Different policies trained with the same algorithm may thus lead to $\mathcal{S} \to \mathcal{A}$ mappings with very different decision boundaries. Transferring targeted examples may not be easy. In order to attack a policy $\pi$, training imitation models like the one from \citet{hester2018deep} to obtain proxy policies that are both (i) solving the same task as $\pi$ and (ii) similar mappings $\mathcal{S} \to \mathcal{A}$ may be the key to compute transferable adversarial examples.
\clearpage

\bibliographystyle{ACM-Reference-Format}  %
\bibliography{bibliography}  %

\end{document}